\documentclass[sigconf]{acmart}
\usepackage{tcolorbox}
\usepackage{colortbl}
\usepackage[normalem]{ulem}
\useunder{\uline}{\ul}{}
\usepackage{multirow}
\usepackage{subcaption}
\usepackage{enumitem}
\usepackage[ruled,vlined]{algorithm2e}

\AtBeginDocument{%
  }

\setcopyright{acmlicensed}
\copyrightyear{2024}
\acmYear{2024}
\acmDOI{XXXXXXX.XXXXXXX}

\acmConference[xx]{Make sure to enter the correct
  conference title from your rights confirmation email}{xxx}{2025}
\acmISBN{978-1-4503-XXXX-X/18/06}
\begin{document}

\title{Bootstrapping your behavior: a new pretraining strategy for user behavior sequence data}


% \affiliation{%

\author{Weichang Wu$^1$, Xiaolu Zhang$^1$, Jun Zhou$^1$, Yuchen Li$^2$, Wenwen Xia$^{3}$}
\authornote{Corresponding author.}
\affiliation{
\institution{$^1$Ant Group, China; $^2$Singapore Management University, Singapore}
  \institution{$^3$Soochow University, China}
  % \city{Hangzhou}
  \country{China}
  }
\email{{jiuyue.wwc,yueyin.zxl,jun.zhoujun}@antgroup.com, yuchenli@smu.edu.sg, wwxia@suda.edu.cn}

% $^1$Ant Group, $^2$Singapore Management University

% \author{Yuchen Li}
% \affiliation{%
%   \institution{Singapore Management University}
%   \country{Singapore}}
% \email{yuchenli@smu.edu.sg}

% \affiliation{%

% \affiliation{%

% \affiliation{%

% \affiliation{%

% \affiliation{%

% \affiliation{%

\renewcommand{\shortauthors}{Xia et al.}

\newcommand{\model}{{\text{BYB}}}
\newcommand{\xww}[1]{\textcolor{blue}{#1}}
\newcommand{\lyc}[1]{\textbf{\textcolor{brown}{[Q: #1]}}}

\newcommand{\myp}[1]{\noindent\underline{\textbf{#1}.}}

\begin{abstract}
User Behavior Sequence (UBS) modeling is crucial in industrial applications. As data scale and task diversity grow, UBS pretraining methods have become increasingly pivotal. 
State-of-the-art UBS pretraining methods rely on predicting behavior distributions.
The key step in these methods is constructing a selected behavior vocabulary. However, this manual step is labor-intensive
and prone to bias. The limitation of vocabulary capacity also directly affects models' generalization ability.
In this paper, we introduce Bootstrapping Your Behavior (\model{}), a novel UBS pretraining strategy that predicts an automatically constructed supervision embedding summarizing all behaviors' information within a future time window, eliminating the manual behavior vocabulary selection.
In implementation, we incorporate a student-teacher encoder scheme to construct the pretraining supervision effectively.
Experiments on two real-world industrial datasets and eight downstream tasks demonstrate that \model{} achieves an average improvement of 3.9\% in AUC and 98.9\% in training throughput. Notably, the model exhibits meaningful attention patterns and cluster representations during pretraining without any label supervision.
In our online deployment over two months, the pretrained model improves the KS by about 2.7\% and 7.1\% over the baseline model for two financial overdue risk prediction tasks in the Alipay mobile application, which reduces bad debt risk by millions of dollars for Ant group.
\end{abstract}

\maketitle

\section{Introduction}\label{sec:intro}
A user behavior sequence (UBS) captures user actions over time in contexts such as mobile apps or online e-commerce platforms. 
UBS is fundamental to a variety of downstream applications, such as user personas construction, recommendation systems, and financial risk management \cite{APP-e-commerce, UBS-social-network, UBSpersonas, UBSFraudDetection, UBSAbnormalSeller}. 
Traditional UBS models are generally trained using supervised learning, relying on predefined labels for each task \cite{pi2019practice,chen2018sequential}.
As the industry's data volume and number of tasks increase, pretraining methods for UBS data become increasingly important to support diverse tasks promptly and effectively.
For example, various financial products like Huabei and Jiebei in Ant Group involve numerous overdue risk prediction tasks. As these products continue to evolve, having a pre-trained model capable of simultaneously handling these diverse and interrelated tasks is essential for minimizing bad debt risks.

% For example, numerous overdue risk prediction tasks exist for different financial products in Ant group, such as Huabei and Jiebei. It is crucial to have a pretrained model that can support these diverse and interrelated tasks simultaneously to reduce bad debt risks.

\begin{figure*}[ht]
  \includegraphics[width=\textwidth]{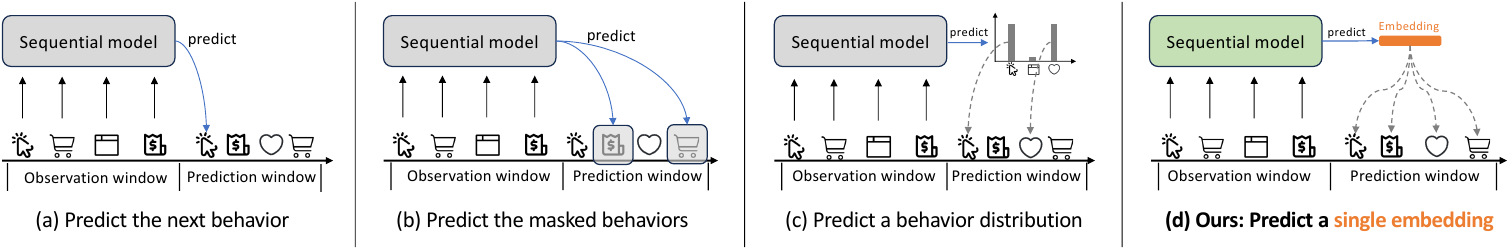}
  % \vspace{-0.5cm}
  \caption{Conceptual comparison between our pretraining strategy and previous pretraining strategies for UBS data.
  }
  \label{fig:fremework}
  % \vspace{-0.5cm}
\end{figure*}

Pretraining has emerged as a powerful technique for natural language processing (NLP) in recent years \cite{GPT3,GPT4,Gemini}.
Inspired by the success of pretraining in NLP, many UBS pretraining methods have adopted similar approaches, by either predicting the next behavior \cite{PTUM} (see \underline{Fig. \ref{fig:fremework}(a)}), or masked behaviors \cite{ShopperBERT,Userbert,RESETBERT4Rec} (see \underline{Fig. \ref{fig:fremework}(b)}).
However, directly applying pretraining strategies from the NLP domain to UBS data does not yield satisfactory performance on downstream tasks, because UBS has unique characteristics—such as local randomness and global orderliness—that differ fundamentally from the structured nature of language. For example, in a mobile financial app, a user might switch between stock and fund sections in any order, with no significant impact on the sequence representation (local randomness). Over a longer period, user behavior might shift from higher-risk activities (stocks) to lower-risk ones (fixed deposits), demonstrating long-term trends (global orderliness).

Hence, recent works for UBS pretraining aim to predict user behavior distributions \cite{MSDP,SUMN}, i.e., pre-defining a behavior vocabulary and predicting whether each behavior in the vocabulary will happen in a future time window (see \underline{Fig. \ref{fig:fremework}(c)}).
Because user behavior distributions tend to be more stable and informative, pretraining models on this strategy lead to state-of-the-art (SOTA) performance. However, a crucial limitation still exists for these methods, i.e., limited behavior vocabulary.

SOTA UBS pretraining methods rely on a manually selected subset of behaviors (limited behavior vocabulary).
The selected behavior vocabulary must be relevant to downstream tasks.
Because the model only focuses on these selected behaviors in pretraining, the quality of behavior vocabulary directly affects model performance.
This manual step can be both labor-intensive and prone to human bias. 
Moreover, the limited vocabulary often overlooks long-tail behaviors, which are crucial for model generalization. 
Expanding the vocabulary to include more behaviors is computationally expensive and often impractical, as behavior space can be vast—ranging from tens of thousands to millions of unique behaviors \cite{TmallDataset1, CLUE}.

% \noindent\underline{\textit{(2) Neglect of local randomness.}} Existing methods directly extract representations on the raw sequence at the behavior level \cite{MSDP,CLUE,SUMN}, but this design ignores the unique characteristic of local randomness in UBS. Neglecting such unique traits introduces noise into the model and largely decreases training efficiency.

To this end, in this paper, we propose a novel pretraining strategy for UBS data, i.e., \underline{B}ootstrapping \underline{Y}our \underline{B}ehavior (\model{}), and a practical implementation, to address the crucial challenge of behavior vocabulary in SOTA UBS pretraining methods.

\noindent\underline{\textit{Pretraining Strategy.}} 
Unlike existing methods that predict individual behaviors based on a limited vocabulary, 
\model{} eliminates the need for manually curated behavior vocabularies by generating a supervision embedding that summarizes all behaviors within a future time window(see \underline{Fig. \ref{fig:fremework}(d)}). This approach not only removes the dependency on manual behavior vocabularies but also captures both frequent and long-tail behaviors smoothly.

\noindent\underline{\textit{Practical Implementation.}} 
In industrial applications, behaviors are typically represented by discrete integer IDs (e.g., 1 for 'stocks', 2 for 'funds') rather than semantically rich embeddings. While previous UBS pretraining methods predict the occurrence of behaviors—making it straightforward to construct discrete pretraining labels from raw IDs—\model{} faces the challenge of generating an effective \textit{embedding} as supervision from these raw IDs that encapsulates all behaviors' information in a future time window. To address this, we introduce a student-teacher behavior encoder scheme, as illustrated in Fig. \ref{fig:framework detailed}. 
The student encoder converts raw behaviors in the observation window $[0, T]$ into embeddings for sequence representation extraction, while the teacher encoder processes behaviors in the prediction window $[T, T+\Delta T_2]$ to construct the supervision embedding. The pretraining objective is to predict this supervision embedding using a predictor module, with the extracted sequence representation being applied in downstream tasks post-pretraining. This innovative approach overcomes the limitations of traditional methods and provides a robust framework for UBS pretraining in real-world industrial applications.

We summarize our contributions in this paper as follows:
\begin{itemize}[leftmargin=*]
\item We propose a novel pretraining strategy for general UBS data (Sec. \ref{sec: overall strategy}) that eliminates the labor-intensive behavior vocabulary construction common in SOTA methods.
\item We implement this strategy in \model{} (Sec. \ref{sec: implementation} - Sec. \ref{sec:pretrain objective and upd}) by incorporating the innovative student-teacher encoder design, specifically designed for industrial UBS data with a focus on both efficiency and effectiveness.
\item We conduct extensive experiments in offline datasets. \model{} improves AUC by 3.9\% over SOTA baselines and boosts training throughput by 98.9\% on two industrial datasets with eight tasks (Sec. \ref{sec: exp setup} - Sec. \ref{sec: exp efficiency}). We also conducted a thorough analysis of the model's intrinsic patterns and extracted UBS embeddings (Sec. \ref{sec: exp scaling} - Sec. \ref{sec: exp visualuzation}).
\item We deploy and evaluate the model in online environments over two months. The model improves the KS by 2.7\% and 7.1\% for two financial overdue risk prediction tasks, significantly reducing bad debt risk by millions of dollars for Ant group (Sec. \ref{sec: exp online exp}).
\end{itemize}

% \vspace{-1em}

% Experiments on two real-world industrial datasets across eight downstream tasks demonstrate that \model{} significantly outperforms state-of-the-art models.
% \model{} achieves an average AUC improvement of 3.9\% over state-of-the-art baselines, with a 98.9\% increase in training throughput.
% Additionally, the pretrained model is capable of learning meaningful attention patterns such as periodicity and time-decay effect, and the pretrained representations without any label supervision exhibit clusters when colored with downstream labels.
% In online deployment over two months, the pretrained model improves the KS by about 2.7\% and 7.1\% for two financial overdue risk prediction tasks over the baseline in the Alipay application.

\section{Related work}
\subsection{User behavior sequence modeling}\label{sec:related work}
We categorize UBS pretraining methods into three main types, i.e.,  behavior prediction, contrastive learning, and behavior distribution prediction.

Most works fall into the \textit{behavior prediction} category. They mainly follow the pretraining strategy in the NLP domain. These approaches can be classified into two main types: masked behavior modeling and next behavior prediction.
In implementation, they either predict randomly masked behaviors in the UBS \cite{PTUM,RESETBERT4Rec,ShopperBERT,Userbert} or predict the next one or several behaviors using historical behaviors \cite{PTUM,Pinnerformer}.

\textit{Contrastive learning} methods primarily focus on constructing positive and negative UBS pairs, e.g., transforming UBS using data augmentation \cite{CLUE,RESETBERT4Rec} or selecting subsequences from the same user to establish positive pairs \cite{Userbert}. Negative pairs are typically formed by randomly sampling sequences from different users.

\textit{Behavior distribution prediction} methods aim to predict the occurrence of a selected subset of behaviors.
They either predict the number of behaviors in the future \cite{SUMN} or the presence of behaviors \cite{MSDP}.
The state-of-the-art method MSDP predicts behavior occurrence within a vocabulary of 200 manually selected behaviors \cite{MSDP}.
However, these methods require manually crafted behavior vocabularies, which are erratic and may hinder generalization abilities. In our work, we learn to predict the automatically constructed supervision embedding that includes all behaviors' information in a future time window. 

\subsection{Self-supervised pretraining}
Self-supervised pretraining has emerged as a powerful technique in NLP \cite{Bert, GPT1, GPT2, GPT3, GPT4, Gemini, T5} and computer vision (CV) \cite{DINO, Dinov2, JEPA, BYOL, iBOT, MAE} domains.
In early studies, masked token prediction and sentence contrastive tasks proved effective as pretext tasks for language data pretraing \cite{Bert}. Language modeling, i.e., predicting the next token, has since become the \textit{de facto} standard revolutionizing large language model pretraining \cite{GPT1, GPT2, T5, GPT3, GPT4, Gemini}.

For vision data, contrastive and reconstruction objectives are commonly employed pretraining strategies. Contrastive learning aims to create positive and negative pairs of images and predict binary signals \cite{MOCO, MOCOv2}, cluster assignment signals \cite{DeepCluster, SwAV}, or representation signals \cite{SimCLR, BYOL, SimSiam, DINO}. Reconstruction objectives involve masking patches of images and training the model to predict these masked areas either in the original image space \cite{MAE, videoMAE} or in a representation space \cite{iBOT, Dinov2, JEPA}.
Multimodal models also pretrain on image-text pairs collectively \cite{CLIP, SimVLM}. 

However, as stated in Sec. 1, UBS data presents unique traits and challenges, making it nontrivial to apply pretraining methods designed for language and vision data directly. Instead, our work is specifically tailored for self-supervised pretraining on UBS data.
% Please refer to Appendix \ref{sec:related work detailed} for an in-depth discussion of related work.

% \xww{
% Self-supervised pretraining has emerged as a powerful technique in natural language processing (NLP) \cite{Bert,GPT1,GPT2,GPT3,GPT4,Gemini,T5} and computer vision (CV) \cite{DINO,Dinov2,JEPA,BYOL,iBOT,MAE} domains.
% Masked token prediction and sentence contrastive tasks are effective pretext tasks for language models in early works \cite{Bert}.
% Recently, language modeling, i.e., predicting the next token, has revolutionized large language model training and becomes the \textit{de facto} standard \cite{GPT1,GPT2,T5,GPT3,GPT4,Gemini}.
% For vision data, contrastive and reconstruction objectives are two commonly adopted strategies. 
% Contrastive learning aims to construct positive pairs and negative pairs of images and predict the positive pairs against negative pairs. 
% Predict binary signals \cite{MOCO,MOCOv2}, cluster assignment signals \cite{DeepCluster,SwAV}, or representation signals \cite{SimCLR,BYOL,SimSiam,DINO}.
% The reconstruction objective mainly masks out some patches of the images and forces the model to predict masked areas either in the original image space \cite{MAE,videoMAE}, or in a representation space \cite{iBOT,Dinov2,JEPA}.
% There are also some multi-modal model pretrain on image-text pairs collectively \cite{CLIP,SimVLM}.
% However, UBS data poses unique characteristics. It is nontrivial to directly utilize pretraining methods for language and vision data. Our work is specifically tailored for pretraining on UBS data.
% }

\section{Preliminaries}
\myp{Data format}
We define a UBS sample $s$ with length $n$ as
\begin{equation}
    s = \{(x_1, t_1), (x_2, t_2), \cdots, (x_n, t_n)\}
\end{equation}
where $x_j$ indicates the $j$-th behavior, and $t_i$ is the logged timestamp for behavior $x_i$.
% $x_1$ to $x_n$ are ordered chronologically during time period $[0, T]$ with $x_n$ the freshest. 
For each behavior $x$, we assume $x$ could contain multiple IDs, i.e., $x=(i_1, \cdots, i_{m})$ with $m$ the number of IDs. We assume the $m$ can differ across different behaviors, the IDs are non-negative integers, and the maximum ID among all possible behaviors is $I$, i.e., $i\in \mathbb{N}_{\leq I}$.

\begin{tcolorbox}[colback=gray!5, colframe=black,boxrule=0.5pt,boxsep=-2pt]
\textbf{Example.}
Assuming each section in a mobile application can be encoded by two positional codes, we can represent a browsing behavior on a specific section with three IDs, e.g., $x=(1, 2, 10)$, where $(1, 2)$ indicates the section's position code and 10 represents that the staying time on the section is 10s. The logged time $t$ could be 2024-01-01,12:00:00.
\end{tcolorbox}

\myp{Problem formulation}
The UBS model aims to take $s$ as input and output an embedding  $\mathbf{E} \in \mathbb{R}^{d}$, $d$ is the dimension, i.e., $\mathbf{E}=f(s)$. In pretraining, $f$ is learned using a large amount of unlabeled data $\{s_1, s_2, \cdots\}$.
Once $f$ is learned, the extracted representation $\mathbf{E}$ is expected to support diverse downstream tasks related to user behavior sequences, such as user profiling or financial risk prediction.

% \lyc{If space permits, describe some downstream tasks.}

\section{Methodology}

\subsection{Overall pretraining strategy}\label{sec: overall strategy}
\begin{figure}[t]
  \includegraphics[width=\linewidth]{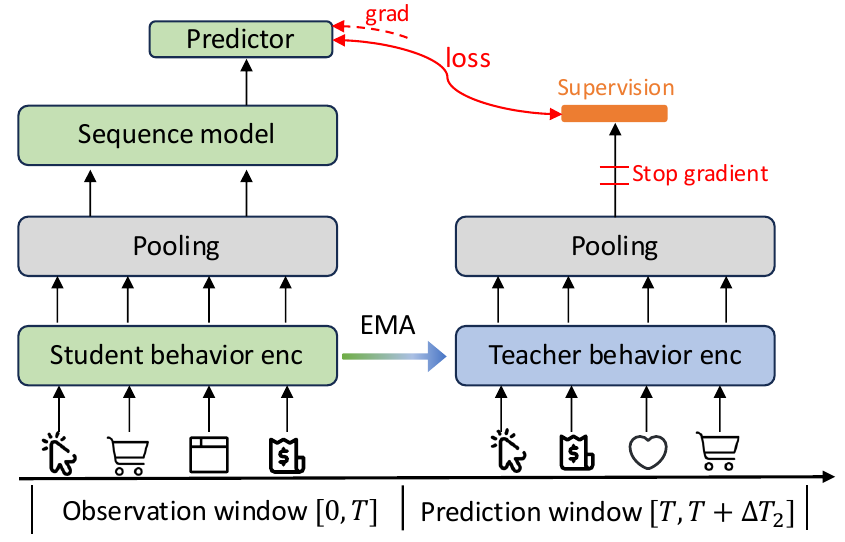}
  % \vspace{-0.5cm}
  \caption{Detailed implementation of \model{}.
  }
  \label{fig:framework detailed}
  % \vspace{-0.5cm}
\end{figure}

As shown in Fig. \ref{fig:fremework} (d), the high-level idea of our pretraining strategy for UBS data is to predict an embedding that includes all behaviors' information in the prediction window, based on the observed behavior sequence in the observation window.
We illustrate the implemented architecture of our pretraining strategy in Fig. \ref{fig:framework detailed}.

Assume the observation time window is $[0, T]$ and the prediction window is $[T, T+\Delta T_2]$, where the $\Delta T_2$ is a predefined prediction time window size. We break down the pretraining procedure into the following three parts.

\noindent$\blacksquare$Firstly  (Fig. \ref{fig:framework detailed} left, from bottom to top), we utilize the student behavior encoder to convert behaviors in the observation window $[0, T]$ into behavior embeddings.
We conduct \textit{behavior pooling} on the behavior embedding sequence to utilize the local randomness and shorten the sequence, which also accelerates computation significantly.
A sequence model then takes the pooled behavior embedding sequence as input and outputs the sequence representation. 
We utilize a predictor module to predict the supervision embedding based on the extracted sequence representation.

\noindent$\blacksquare$Secondly (Fig. \ref{fig:framework detailed} right, from bottom to top), the teacher behavior encoder converts behaviors in the prediction time window $[T, T+\Delta T_2]$ into behavior embeddings.
We conduct a similar pooling operation to construct a single embedding as our pretraining supervision. 
Note that we stop the gradients of the supervision embedding because we expect the teacher behavior encoder to only provide the supervision, instead of optimizing its parameters.

\noindent$\blacksquare$Finally (Fig. \ref{fig:framework detailed} middle, from top to bottom), the training objective is to minimize the discrepancy between the predictor output and the supervision embedding. 
We only update the predictor, the sequence model, and the student behavior encoder using gradient backpropagation.
Since the supervision has no gradients, we update the teacher behavior encoder from the student behavior encoder with an exponential moving average (EMA) strategy, for generating stable supervision signals during pretraining.
The pseudocode is shown in Algorithm \ref{algo: pretraining} in the Appendix.

\subsection{Detailed implementation}\label{sec: implementation}
In this section, we introduce our detailed model implementation for \model{}.
Note that \textbf{the concrete choice of model components in our pretraining framework can be flexible}, we implement our pretraining strategy with the most intuitive and efficient solution for practical industrial deployments.

\myp{(Student and teacher) Behavior encoder} The behavior encoder encodes each behavior into an embedding. Recall that a behavior $x$ is defined as $x=(i_1, \cdots, i_m)$.
In our implementation, we transform each integer ID into an embedding and then merge these embeddings. The behavior encoder consists of two submodules, i.e., the \textit{embedding layer} and the \textit{merge lager}.

The \textit{embedding layer} directly maps each discrete integer ID of a behavior $x$ into a $d$-dimensional embedding, i.e., 
\begin{equation}
    e_j = \texttt{Embedding}(i_j), \quad j \in \{1, \cdots, m\}
    \label{eq:embedding}
\end{equation}
with $e_j \in \mathbb{R}^{d}$. The layer is instantiated as an embedding table.

The \textit{merge layer} merges the $m$ embeddings of one behavior into one embedding. Note that flexible choices exist for implementing this component.
For efficiency consideration, we implement the merge layer as a weighted sum of linearly transformed embeddings, 
\begin{equation}
\begin{aligned}
    \tilde{e}_j &= \mathbf{W}_1 (e_j + p_j) + \textbf{b}_1, 
    \omega_j = \sigma(\mathbf{W}_2 (e_j + p_j) + \textbf{b}_2) \\
    \mathbf{e}_x &= \sum_{j=1}^{m} \omega_j \cdot \tilde{e}_j
\end{aligned}
\label{eq:merge layer}
\end{equation}
where $\mathbf{W}_1 \in \mathbb{R}^{d\times d}$, $\mathbf{b}_1 \in \mathbb{R}^d$, $\mathbf{W}_2 \in \mathbb{R}^{1\times d}$, $\mathbf{b}_2 \in \mathbb{R}^d$, $\mathbf{e}_x \in \mathbb{R}^{d}$, and $\sigma$ is the Sigmoid function. We add position encodings $p_j$ on each ID embedding to capture the order information. Note that we use $\mathbf{e}_x$ to represent the behavior embedding of behavior $x$. 

\myp{Behavior pooling} UBS exhibits local randomness in its raw form, and the sequence length can be long, e.g., thousands of behaviors. We conduct the behavior pooling on the behavior embedding sequence to alleviate the randomness and accelerate computation.
% before taking it into the sequence model.

We conduct behavior pooling based on time, e.g., days, since behaviors are generally logged with timestamps, e.g., 2024-01-01, 12:00:00, in industrial applications.
We denote the pooling time window as $\Delta T_1$. 
The behavior pooling is as follows.
\begin{equation}
    \bar{\mathbf{e}}_k = \texttt{BehPool}(
    \{
    \mathbf{e}_j | t_j \in [k\Delta T_1, (k+1)\Delta T_1] 
    \}
    ), k \in \mathbb{N}_{\leq \frac{T}{\Delta T_1}}
    \label{eq:beh pooling}
\end{equation}
Here we use $\mathbf{e}_j$ to indicate the $j$-th behavior's embedding in $s$, and $\bar{\mathbf{e}}_k$ to represent the pooled embedding of the $k$-th window (assume $T$ is divisible by $\Delta T_1$).
In our implementation, \texttt{BehPool} is a mean operator. Although other operators (max, sum, or even RNNs) may also be eligible, we found that the mean operator works well for most tasks in preliminary experiments and is also very efficient.

We set the stride equal to the window size $\Delta T_1$.
After behavior pooling, the sequence length is shortened from $n$ to 
$T/\Delta T_1$.
The shortened sequence could accelerate computation significantly since the complexity of widely used self-attention-based models scales quadratically w.r.t sequence length. 

\begin{tcolorbox}[colback=gray!5, colframe=black,boxrule=0.5pt,boxsep=-2pt]
\textbf{Clarification on pooling window and stride choice.}
In industrial applications, user behaviors generally exhibit stable patterns at the one-day granularity, e.g., people's daily habits and usage in a mobile app can be steady and change slowly.
By default, we select one day as the pooling window size and stride.
Our preliminary experiments show that one day works very well for most tasks.
\end{tcolorbox}

\myp{Sequence model and predictor}
The sequence model \texttt{SeqModel} aims to extract a sequence representation $\mathbf{E}$ from the pooled behavior sequence. The representation $\mathbf{E}$ is calculated as in Eq. \ref{eq:seqmodel}.
\begin{equation}
    \mathbf{E} = \texttt{SeqModel} ([\bar{\mathbf{e}}_{1}, \bar{\mathbf{e}}_{2}, \cdots, \bar{\mathbf{e}}_\frac{T}{\Delta T_1}])
    \label{eq:seqmodel}
\end{equation}
where $\mathbf{E} \in \mathbb{R}^d$.
Note that $\frac{T}{\Delta T_1}$ is the new sequence embedding length.
We instantiate \texttt{SeqModel} with a Transformer encoder \cite{Transformer}. Positional encoding is added to the $[\bar{\mathbf{e}}_j]$ sequence in the \texttt{SeqModel}.
The last embedding of \texttt{SeqModel}'s last layer is regarded as $\mathbf{E}$.

In pretraining, we expect that the sequence representation $\mathbf{E}$ can predict the supervision embedding that includes all behaviors' information in the prediction window. 
Hence, we use a predictor module that takes $\mathbf{E}$ as input and outputs a prediction $\hat{\mathbf{E}}$ for the supervision embedding, i.e., 
\begin{equation}
    \hat{\mathbf{E}} = \texttt{Predictor} (\mathbf{E})
    \label{eq:predictor}
\end{equation}
where $\hat{\mathbf{E}} \in \mathbb{R}^d$ is the prediction and will be utilized to calculate the loss with the supervision embedding.
The \texttt{Predictor} is implemented with a two-layer MLP.
Note that the predictor module is only used during pretraining and is discarded in finetuning.

\myp{Supervision embedding for pretraining}\label{sec:supervision signal}
In a nutshell, the supervision is a single embedding and is calculated using behavior embeddings from the teacher behavior encoder, as shown in Fig. \ref{fig:framework detailed} (right).
\textit{The teacher behavior encoder possesses an identical architecture to the student behavior encoder (rf. Eq. \ref{eq:embedding} and Eq. \ref{eq:merge layer}). However, they take different behaviors as input during pretraining.}

We denote the prediction window size as $\Delta T_2$, i.e., the time horizon we hope the model predicts for the future.
\textit{Firstly}, all behaviors in the prediction window $[T, T+\Delta T_2]$ will be converted into behavior embeddings \textit{using the teacher behavior encoder}.
\textit{Secondly}, since we aim to construct a single embedding as supervision, we conduct behavior pooling to calculate the supervision embedding $\bar{\mathbf{E}}$.
\begin{equation}
    \bar{\mathbf{E}} = \texttt{BehPool}(
    \{
    \mathbf{e}_j | t_j \in [T, T+\Delta T_2] 
    \}
    )
    \label{eq:target pooling}
\end{equation}
where $\bar{\mathbf{E}} \in \mathbb{R}^d$.
We instantiate the \texttt{BehPool} in Eq. \ref{eq:target pooling} as a mean operator, analogous to Eq. \ref{eq:beh pooling}.

\textit{Finally, and importantly}, since we expect the teacher behavior encoder to only provide stable supervision signal in pretraining, instead of optimizing its parameters, \textit{we conduct a \textit{stop-gradient} (\texttt{stopgrad}) operation on the supervision} $\bar{\mathbf{E}}$ as follows.
\begin{equation}
    \bar{\mathbf{E}} = \texttt{stopgrad}(\bar{\mathbf{E}})
    \label{eq:stopgrad}
\end{equation}
% $\bar{\mathbf{E}}$ has no gradients and hence the teacher behavior encoder will not be updated by gradient back-propagation. 
We elaborate on the parameter updating strategy in Sec. \ref{sec:pretrain objective and upd}.

% \begin{tcolorbox}[colback=gray!5, colframe=black,boxrule=0.5pt,boxsep=-2pt]
% \textbf{Clarification on prediction window choice.}
% Based on industrial applications, the prediction window is generally set to one to several days. In our preliminary experiments, one day (default setting) works well for most downstream tasks. We also have an ablation experiment to investigate the prediction window size effect in Sec. \ref{sec:exp pred window size}.
% \end{tcolorbox}

The prediction window is generally set to one to several days. One day (default setting) works well for most downstream tasks. We also have an ablation experiment to investigate the prediction window size effect in Sec. \ref{sec:exp pred window size}.

\subsection{Pretraining objective and updating strategy}\label{sec:pretrain objective and upd}
% In this section, we introduce the pretraining objective and parameter updating strategy of \model{}.
\myp{Pretraining objective}
We resort to some similarity metrics to measure the discrepancy between the predicted embedding $\hat{\mathbf{E}}$ (rf. Eq. \ref{eq:predictor}) and the supervision embedding $\bar{\mathbf{E}}$ (rf. Eq. \ref{eq:stopgrad}).
We write the pretraining objective in a causal form as follows.
\begin{equation}
    L = \sum_{k=1}^{T/\Delta T_1} l(\hat{\mathbf{E}}_k, \bar{\mathbf{E}}_k)
    \label{eq:causal loss}
\end{equation}
$l$ is instantiated by a specific similarity measure. The loss $L$ in Eq. \ref{eq:causal loss} calculates the discrepancy between $\hat{\mathbf{E}}$ and $\bar{\mathbf{E}}$ for all observation sub-windows $[0, k\Delta T_1]$  and corresponding prediction windows $[k\Delta T_1, k\Delta T_1 + \Delta T_2]$, where $k=1, 2, \cdots, T/\Delta T_1$.

The similarity measure is not unique and can be flexible for different scenarios.
We mainly adopt two implementations, i.e., the cross-entropy and the mean square error.
For the cross-entropy loss $l_{\texttt{ce}}$, we conduct a Softmax operator to normalize embeddings and then calculate the cross-entropy between the two embeddings, i.e., 
% (also referred to as sharpening in the literature \cite{DINO}) 
\begin{equation}
\begin{aligned}
\texttt{Softmax}(\mathbf{x})_i &= \frac{\exp{(\mathbf{x}_i / \tau)}}{\sum_{j=1}^{d} \exp{(\mathbf{x}_j / \tau)} } \\
l_{\texttt{ce}}(\mathbf{x}, \mathbf{y}) &= \texttt{CE}(\texttt{Softmax}(\mathbf{x}), \texttt{Softmax}(\mathbf{y})) \\
\end{aligned}
\end{equation}
where $\tau$ is a temperature, e.g., 0.1, and $\texttt{CE}(\mathbf{x},\mathbf{y}) = -\sum_{i=1}^{d}\mathbf{x}_i\log \mathbf{y}_i $.
% \begin{equation}
%     l_{\texttt{ce}}(\mathbf{x}, \mathbf{y}) = \texttt{CE}(\texttt{Softmax}(\mathbf{x}), \texttt{Softmax}(\mathbf{y}))
% where $\texttt{CE}(\mathbf{x},\mathbf{y}) = -\sum_{i=1}^{d}\mathbf{x}_i\log \mathbf{y}_i $,  indicating the cross-entropy. 
The mean square error loss $l_{\texttt{mse}}$ is defined as follows.
% \begin{equation}
%     l_{\texttt{mse}}(\mathbf{x}, \mathbf{y}) = \frac{1}{d} \sum_{l=1}^{d} (\mathbf{x}_i - \mathbf{y}_i)^2
% \end{equation}
\begin{equation}
    l_{\texttt{mse}}(\mathbf{x}, \mathbf{y}) = \frac{1}{d} \Vert \mathbf{x} - \mathbf{y} \Vert_2^2
\end{equation}
In experiments, we find that both similarity measures have their pros and cons. By default, we adopt the cross-entropy loss. We will compare the effect of the two measures in Sec. \ref{sec:exp}.

\myp{Updating strategy}
As shown in Fig. \ref{fig:framework detailed} (middle top), \textit{we update the student behavior encoder, the sequence model, and the predictor using gradient backpropagation computed from the loss}.
% the supervision embedding $\bar{\mathbf{E}}$ is constructed from the output of the teacher behavior encoder, and we have stop its gradient during pretraining (rf. Eq. \ref{eq:stopgrad}).
% We \textit{stop the gradients of the supervision signal} in pretraining since we expect the teacher behavior encoder to provide stable supervision signals instead of optimizing its parameters.

For the teacher behavior encoder, it will not be updated using gradients since we stop the gradient of $\bar{\mathbf{E}}$ (rf. Eq. \ref{eq:stopgrad}).
\textit{We update it using an EMA strategy from the student behavior encoder}, to obtain stable and informative supervision, inspired by deep RL \cite{DQN,A3C}. The EMA updates the teacher behavior encoder's parameters as follows:
\begin{equation}
    \theta^t_{i+1} \leftarrow m \theta^t_{i} + (1-m) \theta^s_{i+1}
\end{equation}
where $\theta^t$ and $\theta^s$ indicate the teacher and the student behavior encoders' parameters, respectively, and $m$ is a momentum hyperparameter (0.995 as default in our implementation).

\section{Experiments}\label{sec:exp}
% Overall experimental design.

\subsection{Experimental setup}\label{sec: exp setup}
\subsubsection{Datasets.}

We utilize two large real-world industrial datasets and eight business tasks: the Tmall dataset and the Alipay Mobile Application dataset. The public Tmall dataset captures user-item interactions \cite{TmallDataset1, TmallDataset2} on the Tmall platform \footnote{https://tianchi.aliyun.com/dataset/140281}, while the Alipay Mobile Application logs user behaviors within the Alipay application and is private to Ant Group \footnote{https://www.antgroup.com/en} for the current time. The Tmall dataset comprises 8,000,000 distinct behavior IDs, while the Mobile Application has 30,000 IDs. Both datasets are divided into three subsets each for pretraining, finetuning, and testing, as detailed in Tab. \ref{tab:dataset statistics}. 
The Tmall dataset contains about 3.9 million samples while the Mobile dataset contains about 10 million samples. 
We employ four downstream tasks for each dataset. Specifically, the Tmall tasks involve predicting users'  most interested item category in the future 5, 10, 15, and 30 days, all framed as 90-class classification problems (tasks named \texttt{5-day}, \texttt{10-day}, \texttt{15-day}, and \texttt{30-day}). The tasks for the Mobile Application dataset focus on predicting users' activities within the app, including whether users will be active, login, and unbind their accounts within the next 30 or 90 days (tasks named \texttt{Active-30d}, \texttt{Login-90d}, \texttt{Unbind-30d}, and \texttt{Unbind-90d}). All tasks are binary classifications. 
Please refer to Appendix \ref{app:datasets} for more dataset and preprocessing details.

% We adopt two datasets for pretraining and downstream finetuning, i.e., Tmall and Mobile application. 

% Tmall is a recommendation dataset recording user-item interactions \cite{TmallDataset1,TmallDataset2}. 
% Mobile application is a log dataset recording user's behaviors in the Alipay \footnote{\url{https://www.alipay.com}} mobile application \cite{MSDP}.

% The number of distinct behavior IDs $I$ is 8000000 and 30000 for Tmall and Mobile application datasets, respectively.

% Each dataset is split into three subsets, i.e., pretraining, finetuning, and testing. The statistics are detained in Tab. \ref{tab:dataset statistics}.

% We adopt four downstream tasks for each dataset. 
% For the Tmall dataset, the tasks are to predict a user's future most interested item category, i.e., y-5d,  y-10d,  y-15d,  y-30d.
% All are 90-class classification tasks.

% For the Mobile application dataset, the tasks are to predict a user's state within the app, i.e., whether he will be active (active-30d, active-90d) and whether he will withdraw his account (unbind-90d, unbind-90d).
% All are binary classification tasks.

% We refer readers to the Appendix \ref{app:datasets} for more details about the datasets, tasks, and our pre-processing and splitting details.

% \usepackage{multirow}
\begin{table}[]
\centering
\caption{Dataset statistics.}
% \vspace{-0.3cm}
\label{tab:dataset statistics}
\resizebox{\linewidth}{!}{%
\begin{tabular}{c|c|cccc}
\hline
Dataset                   & Split    & \# Samples & Max len & Avg len & \# Days \\ \hline
\multirow{3}{*}{Tmall}    & Pretrain & 3.9$\times 10^6$      & 600     & 130.4   & 120  \\
                          & Finetune & 2.6$\times 10^6$      & 579     & 71.6    & 60   \\
                          & Test     & 2.9$\times 10^6$      & 585     & 71.3    & 60   \\ \hline
\multirow{3}{*}{Mob. app} & Pretrain & 10$\times 10^6$       & 8994    & 2129.7  & 180  \\
                          & Finetune & 4.2$\times 10^6$      & 8990    & 2114.5  & 180  \\
                          & Test     & 2.3$\times 10^6$      & 8991    & 2104.3  & 180  \\ \hline
\end{tabular}
}
% \vspace{-0.5cm}
\end{table}

\subsubsection{Baselines}
We compare our method with both pretraining and supervised learning methods.
For pretraining methods, we consider three kinds of baselines, i.e., behavior prediction methods, contrastive learning methods, and behavior distribution prediction methods (rf. Sec. \ref{sec:related work}). The details of baselines are as follows.
\begin{enumerate}[nosep, leftmargin=*]
  \item \textbf{NBP}. We employ next behavior prediction (NBP) \cite{PTUM} as our baseline, which belongs to the behavior prediction category. NBP predict the next behavior during pretraining \cite{PTUM}.
  \item \textbf{MBMv1, MBMv2}. We employ masked behavior modeling \cite{Userbert,PTUM} as our baseline, which also belongs to the behavior prediction category. We consider two variants featuring mask ratios of 0.1 (MBMv1) and 0.2 (MBMv2), respectively.
  \item \textbf{CTS}. We contrast a user's different UBS with random local permutation as positive pairs and different users' UBS as negative pairs (denoted as CTS) \cite{CLUE,Userbert}, which belongs to the contrastive learning category.
  \item \textbf{MSDP}. Multi-scale stochastic distribution prediction (MSDP) is the \textit{state-of-the-art} method for UBS pretraining \cite{MSDP}. MSDP predicts the presence of a small manually selected set of behaviors in a future time window, belonging to the behavior distribution prediction category.
  \item \textbf{Supervised}. We train a Transformer encoder model from scratch using labels (denoted as Supervised). We select Transformer as our supervised model because it has been the \textit{de facto} standard and \textit{is proven to surpass other sequence models (e.g., RNN, LSTM) for sequence tasks} \cite{TransformerVSRNN}. We added this supervised baseline to compare between self-supervised and supervised methods on UBS data.
\end{enumerate}

% Moreover, we also train a model from scratch using labels as our supervised baseline (denoted as Supervised). We compare the performance of all pretraining methods with the supervised model to investigate the performance gap between self-supervised models and supervised ones on UBS data.

% Since this work focuses on pretraining methods for UBS data, we mainly consider commonly adopted pretraining baselines designed for this specific data modality.

% Classic behavior prediction methods include masked behavior modeling (MBM) \cite{Userbert,PTUM} and next behavior prediction (NBP) \cite{PTUM}. 
% For MBM, we consider two variants with different maske ratios (0.1 and 0.2).
% For NBP, we predict the next behavior implementation \cite{PTUM}.
% Besides, we consider another classic strategy, i.e., contrastive learning-based pretraining (CTS) \cite{CLUE,Userbert}.
% We also consider the state-of-the-art behavior distribution prediction-based pretraining method MSDP \cite{MSDP}.

% We also train a model from scratch with labels as a supervised baseline, (Named Supervised).

% We train our \model{} with pretraining. 
% Our model with all parameters finetuned using labels after pretraining is denoted as \model{}(unfreeze).

\vspace{-0.1cm}
\subsubsection{Evaluation protocols}\label{sec:evaluation protocol}
For the Supervised baseline, we directly \textit{train the model's all parameters} using labels from scratch.
For unsupervised baselines, we first pretrain models without labels, then finetune them using downstream labels. During finetuning, we \textit{freeze the pretrained model's parameters and only finetune a new prediction head} for each downstream task. The prediction head takes the extracted sequence representation  $\mathbf{E}$ (rf. Eq. \ref{eq:seqmodel}) as input.

For our method \model{}, we elaborate on the two variants of our method as follows.
\begin{enumerate}[nosep, leftmargin=*]
\item \textbf{\model{}(freeze)}. During finetuning, we \textit{freeze the pretrained model's parameters and only finetune prediction heads}.
\item \textbf{\model{}(unfreeze)}. During finetuning, we \textit{finetune all parameters including both prediction heads and the pretrained sequence encoder and student behavior encoder}.
\end{enumerate}

We compare \model{}(freeze) with NBP, MBMv1, MBMv2, CTS, and MSDP because they all freeze the main encoder's parameters in finetuning. We compare \model{}(unfreeze) with the Supervised baseline because all parameters of the supervised baseline are trained during training.

For all datasets and downstream tasks, we adopt the Area Under the Receiver Operating Characterization Curve (AUROC) metric by default. For the Tmall dataset with a class number larger than two, we compute the AUROC metric using the Macro strategy.

% \usepackage{multirow}
% \usepackage[normalem]{ulem}
\begin{table*}[ht]
\centering
\caption{Performance comparison of all methods on both datasets. We highlighted the best results in bold and underlined the second-best results \textit{among all frozen pretrained models}, excluding the Supervised and the \model{}(unfreeze) model because they have all parameters trained.
We also use bold to denote the best results across all models, i.e., \model{}(unfreeze).
The \model{}(unfreeze) is highlighted with a yellow background, while the Supervised baseline is highlighted with a gray background.}
% \vspace{-0.1cm}
\label{tab:main results}
\begin{tabular}{cc|cccc|cccc}
\hline
\multicolumn{2}{c|}{Dataset}                                                                        & \multicolumn{4}{c|}{Tmall}                                                                                                                                            & \multicolumn{4}{c}{Alipay Mobile Application}                                                                                                                                         \\ \hline
\multicolumn{2}{c|}{Methods}                                                                        & 5-day                                   & 10-day                                  & 15-day                                  & 30-day                                  & Active-30d                              & Login-90d                               & Unbind-30d                              & Unbind-90d                              \\ \hline
\multicolumn{1}{c|}{Supervised learning}                            & \cellcolor[HTML]{D9D5D5}Supervised           & \cellcolor[HTML]{D9D5D5}0.6267          & \cellcolor[HTML]{D9D5D5}0.6239          & \cellcolor[HTML]{D9D5D5}0.6223          & \cellcolor[HTML]{D9D5D5}0.6147          & \cellcolor[HTML]{D9D5D5}0.9218          & \cellcolor[HTML]{D9D5D5}0.9132          & \cellcolor[HTML]{D9D5D5}0.7462          & \cellcolor[HTML]{D9D5D5}0.7215          \\ \cline{1-2}
\multicolumn{1}{c|}{}                                      & MBMv1                                  & 0.4555                                  & 0.4458                                  & 0.4412                                  & 0.4261                                  & 0.6668                                  & 0.6315                                  & 0.5274                                  & 0.5372                                  \\
\multicolumn{1}{c|}{}                                      & MBMv2                                  & 0.4644                                  & 0.4556                                  & 0.4498                                  & 0.4341                                  & 0.6260                                  & 0.5998                                  & 0.5498                                  & 0.5503                                  \\
\multicolumn{1}{c|}{\multirow{-3}{*}{Behavior prediction}} & NBP                                    & 0.3321                                  & 0.3374                                  & 0.3300                                  & 0.3024                                  & 0.8447                                  & 0.7893                                  & 0.6449                                  & 0.6378                                  \\ \cline{1-2}
\multicolumn{1}{c|}{Contrastive}                           & CTS                                    & 0.4394                                  & 0.4274                                  & 0.4226                                  & 0.4060                                  & 0.8032                                  & 0.7815                                  & 0.6607                                  & 0.6509                                  \\ \cline{1-2}
\multicolumn{1}{c|}{Distribution prediction}               & MSDP                                   & {\ul 0.4799}                            & {\ul 0.4686}                            & {\ul 0.4620}                            & {\ul 0.4476}                            & {\ul 0.8893}                            & {\ul 0.8900}                            & \textbf{0.7425}                         & \textbf{0.7226}                         \\ \hline
\multicolumn{1}{c|}{}                                      & \model{}(freeze)                                    & \textbf{0.5779}                         & \textbf{0.5673}                         & \textbf{0.5618}                         & \textbf{0.5456}                         & \textbf{0.9134}                         & \textbf{0.8950}                         & {\ul 0.6823}                            & {\ul 0.6702}                            \\
\multicolumn{1}{c|}{\multirow{-2}{*}{Ours}}                & \cellcolor[HTML]{FFCE93}\model{} (unfreeze) & \cellcolor[HTML]{FFCE93}\textbf{0.6409} & \cellcolor[HTML]{FFCE93}\textbf{0.6381} & \cellcolor[HTML]{FFCE93}\textbf{0.6364} & \cellcolor[HTML]{FFCE93}\textbf{0.6281} & \cellcolor[HTML]{FFCE93}\textbf{0.9238} & \cellcolor[HTML]{FFCE93}\textbf{0.9224} & \cellcolor[HTML]{FFCE93}\textbf{0.7669} & \cellcolor[HTML]{FFCE93}\textbf{0.7405} \\ \hline
\end{tabular}
\end{table*}

% \vspace{-.5cm}
\subsubsection{Implementation details}\label{sec: implementation details}
To ensure a fair comparison, we utilize identical model architectures for all pretraining methods and the supervised learning baseline.
\textit{The following hyperparameters are based on our preliminary experiments and efficiency requirements for deployment.}
The embedding table dimension is set to 128 for both datasets and all methods.
The behavior encoder's hidden dimension is set to 128.
The main sequential model (and the Supervised baseline) is instantiated as a 4-layer Transformer encoder model \cite{Transformer}. 
Each transformer encoder layer has a hidden dimension and a feed-forward dimension of 128. 
The predictor is a two-layer MLP with hidden dimension 128.
After pretraining, the predictor is discarded.

In finetuning, new prediction heads are created for each downstream task, taking the extracted sequence representation $\mathbf{E}$ (rf. Eq. \ref{eq:seqmodel}) as input. Each head is a two-layer MLP with a hidden dimension of 64, and the output dimension is set to the number of classes.
The \model{} and all comparison models are pre-trained for two epochs, with finetuning for one epoch for each task due to the large datasets. 
We set the default behavior pooling window size to one day.

We adopt the Adam optimizer with a learning rate of 1e-4, a weight decay of 1e-4, and a batch size of 512. 
All experiments are carried out on a machine equipped with an NVIDIA A100 GPU with 80GB high bandwidth memory.

% \vspace{-.5cm}
\subsection{Main results}\label{sec: exp main results}
We illustrate the test results of all models in Tab. \ref{tab:main results}.
As clarified in Sec. \ref{sec:evaluation protocol},
We should compare \model{}(freeze) with NBP, MBMv1, MBMv2, CTS, and MSDP. Because for these models, 
we freeze the encoders's parameters and only train prediction heads during finetuning.
Besides, we should compare \model{}(unfreeze) with the Supervised baseline, because all parameters are trained/finetuned for the two models.
Our conclusions are as follows:

% Note that we highlighted the best results in bold and underlined the second-best results \textit{among all frozen pretrained models}, excluding the Supervised and the \model{}(unfreeze) model because they have all parameters trained.
% We also use bold to denote the best results across all models, i.e., \model{}(unfreeze), in Tab. \ref{tab:main results}.
% The \model{}(unfreeze) is highlighted with a yellow background, while the supervised baseline is highlighted with a gray background.

\noindent$\blacksquare$1, Compared with other frozen pretrained models (MBMv1, MBMv2, NBP, CTS, and MSDP), the \model{}(freeze) achieves the best results on six out of eight tasks. The average improvement over the second-best is 3.9\%, verifying the superiority of our pretraining strategy over previous pretraining methods for UBS data.
The MSDP performs the best among all pretraining baselines, verifying the effectiveness of behavior distribution prediction over behavior prediction and contrastive learning for UBS pretraining. The best performance of CTS, NBP, and MBM on eight downstream tasks varies and is not consistent, which is as expected and indicates that simply adopting pretraining methods in the NLP domain is not suitable for UBS data pretraining.

\noindent$\blacksquare$2, The \model{}(unfreeze) performs the best among all methods.
Significantly, compared with the full-parameter trained Supervised baseline, \model{}(unfreeze) surpasses it on all tasks for both datasets and the average improvement is 1.3\%. The general improvement for diverse downstream tasks verifies the generalization ability and effectiveness of our pretraining strategy.

% Moreover, we can find that for all frozen pretrained models including our \model{}, a performance gap still exists with the supervised baseline. We argue that the performance gap is normal and reasonable since pretraining methods do not utilize the downstream label signal and their parameters are frozen during finetuning.
% Instead, we expect our pretrained model to be a proper initialization for downstream tasks. 

% As shown in the last row of Tab. \ref{tab:main results}, the \model{}(unfreeze) performs the best among all methods. Significantly, it surpasses the supervised baseline on all tasks for both datasets and the average improvement is 1.3\%. The general improvement for diverse downstream tasks verifies the generalization ability and effectiveness of our pretraining strategy.

% We can find the \model{}(unfreeze) suppress the \model{} as well.
% Hence we recommend unfreezing the model parameters during downstream fin

\subsection{Computational efficiency}\label{sec: exp efficiency}

% \usepackage[normalem]{ulem}
\begin{table}[]
\centering
\caption{Training throughput (\# samples per second) comparison on both datasets. FT-train indicates finetuning.}
% \vspace{-0.3cm}
\label{tab:train efficiency}
\begin{tabular}{c|cccc}
\hline
Dataset          & \multicolumn{2}{c}{Tmall}       & \multicolumn{2}{c}{Mobile App.}   \\ \hline
                 & Pretrain       & FT-train       & Pretrain       & FT-train        \\ \hline
Sup.  w/ pooling & -              & \textbf{593.9} & -              & 839.6           \\
MBM              & {\ul 419.8}    & 348.1          & 289.2          & {\ul 926.7}     \\
NBP              & 363.5          & 281.6          & 217.6          & 727.0           \\
CTS              & 327.6          & 358.4          & 144.6          & 921.6           \\
MSDP             & 378.8          & 322.5          & {\ul 291.8}          & 916.4           \\
\model{}     & \textbf{568.3} & {\ul 552.9}    & \textbf{765.8} & \textbf{1070.0} \\ \hline
\end{tabular}
% \vspace{-0.5cm}
\end{table}

Training efficiency is an important factor in industrial applications.
Because industrial datasets are generally vast and long training time results in more cost on both GPU hours and energy.
% Since the pretraining dataset is typically enormous, the computational expense associated with pretraining can be significantly burdensome.
% In this part, we compare the efficiency of all methods.
We list the training throughput of all methods in Tab. \ref{tab:train efficiency}. 
Note that we equip the Supervised baseline with the same behavior pooling (denoted as w/ pooling) as in our \model{} to improve stability and efficiency. 
% The best result is in bold and the second best is underlined.

We can find that in the pretraining stage, the \model{} achieves the highest throughput among all baselines on both datasets.
The average pretraining throughput improvement over the second-best is about 98.9\%. We attribute the efficiency superiority to the behavior pooling in our framework. Since the most computational cost comes from the self-attention layers in the sequential model, which has quadratic cost w.r.t the sequence length. 
Our analysis is substantiated by results from the Mobile app dataset, where the \model{}'s pretraining throughput surpasses other baselines by a notable 162\%. 
This is reasonable given that the Mobile app dataset has a significantly longer average UBS length (approximately 2000) compared to the Tmall dataset (around 130) (rf. Tab. \ref{tab:dataset statistics}). By leveraging the one-day based behavior pooling, we maintain a constant pooled sequence length that is substantially shorter than the original length.
% Given the quadratic complexity of self-attention operations, this behavior pooling design markedly enhances the pretraining speed.
For finetuning throughput, the \model{} consistently exceeds all pretraining baselines, matching the supervised baseline that also utilizes behavior pooling.

% For finetuning throughput, the \model{} consistently outperforms all pretraining baselines and is similar to the supervised baseline since we also adopt behavior pooling for the supervised baseline. The improvement percentage is relatively lower than that in the pretraining stage since we only train prediction heads in the finetuning stage, which is reasonable.

% We analyze that the reason for a superior improvement is that 

% We conduct the behavior pooling by days using timestamps, hence the sequence length is constant. 

% For baselines without behavior pooling, the computational complexity is quadratic w.r.t sequence length, resulting in a much lower training speed.

% 1, the pretraining is the fastest among all pretraining baselines.
% 2, the finetuning is also the fastest among all pretraining baselines.
% 3, the finetuning speed is similar to the supervised baseline.

\begin{figure}[t]
    \centering
    \begin{subfigure}[b]{0.48\linewidth}
        \includegraphics[width=\textwidth]{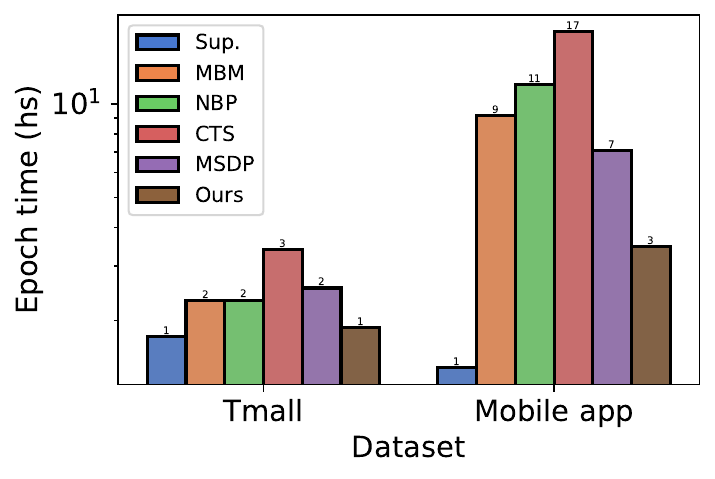}
        \vspace{-0.5cm}
        \caption{Epoch time.} % 如果你不需要caption，可以注释掉这一行
    \end{subfigure}
    % \hfill
    \hspace{-0.2cm}
    \begin{subfigure}[b]{0.48\linewidth}
        \includegraphics[width=\textwidth]{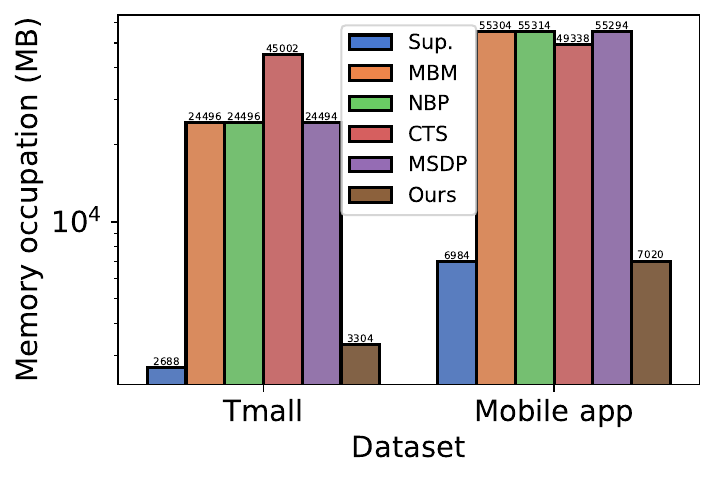}
        \vspace{-0.5cm}
        \caption{Memory occupation.} % 如果你不需要caption，可以注释掉这一行
    \end{subfigure}
    % \vspace{-0.3cm}
    \caption{Epoch time and memory occupation comparison.} % 你可以根据需要添加整个figure的caption
    \label{fig:epoch time and memory}
    % \vspace{-0.4cm}
\end{figure}

We present a comparison of epoch times (time taken for one epoch) and memory utilization in Fig. \ref{fig:epoch time and memory}. 
The shorter epoch time of \model{} than pretraining baselines aligns well with the throughput results in Tab. \ref{tab:train efficiency}. 
Additionally, \model{} demonstrates significantly lower memory usage compared to pretraining baselines, due to the shorter sequence length achieved by behavior pooling.

\subsection{Scaling effect}\label{sec: exp scaling}
\begin{table}[]
\centering
\caption{Performance of scaling the model size.}
% \vspace{-0.3cm}
\label{tab: scaling results}
\resizebox{\linewidth}{!}{%
\begin{tabular}{c|cccc}
\hline
                        & Active-30d      & Login-90d      & Unbind-30d      & Unbind-90d      \\ \hline
\rowcolor[HTML]{EFEFEF} 
\texttt{Base}           & 0.9134          & 0.8950         & 0.6823          & 0.6702          \\
\texttt{Base}$\times2$  & 0.9175          & 0.9014         & 0.6877          & 0.6782          \\
\texttt{Base}$\times4$  & 0.9176          & 0.9029         & 0.6863          & 0.6765          \\
\texttt{Base}$\times8$  & \textbf{0.9178} & {\ul 0.9033}   & \textbf{0.7002} & \textbf{0.6876} \\
\texttt{Base}$\times16$ & {\ul 0.9177}    & \textbf{0.9040} & {\ul 0.6935}    & {\ul 0.6789}    \\ \hline
\end{tabular}
}
\vspace{-0.3cm}
\end{table}

The scaling effect is important when taking into account more data and larger model sizes for pretraining in the industry. To examine the scalability of our method, we increase the size of our default base model (rf. Sec. \ref{sec: implementation details}) by factors of 2, 4, 8, and 16. Detailed configurations are provided in Tab. \ref{tab:scaling model config}. We present the final test results in Tab. \ref{tab: scaling results} on the Mobile application dataset.

As shown in Tab. \ref{tab: scaling results}, it is evident that larger model sizes generally yield better performance (especially on the Unbind-30d and Unbind-90d), verifying the scalability of \model{}. 
Specifically, the \texttt{Base}$\times$8 and \texttt{Base}$\times$16 rank as the top-performing and second-best models, respectively. 
Interestingly, the \texttt{Base}$\times$8 outperforms the larger \texttt{Base}$\times$16, achieving approximately 1.86\% improvement over the \texttt{Base} model in the more challenging Unbind-30d and Unbind-90d tasks. 
We hypothesize that this might be due to architectural differences: \texttt{Base}$\times$8 has a deeper architecture, whereas \texttt{Base}$\times$16 is wider. This suggests that a deeper architecture may be more suitable than a wider one when scaling model size for UBS data pretraining. More details of our scaling experiments settings can be found in Appendix \ref{sec: appendix scaling effect}.

\subsection{Effect of the prediction window size}\label{sec:exp pred window size}

\begin{figure}[ht]
    \centering
    \begin{subfigure}[b]{0.47\linewidth}
        \includegraphics[width=\textwidth]{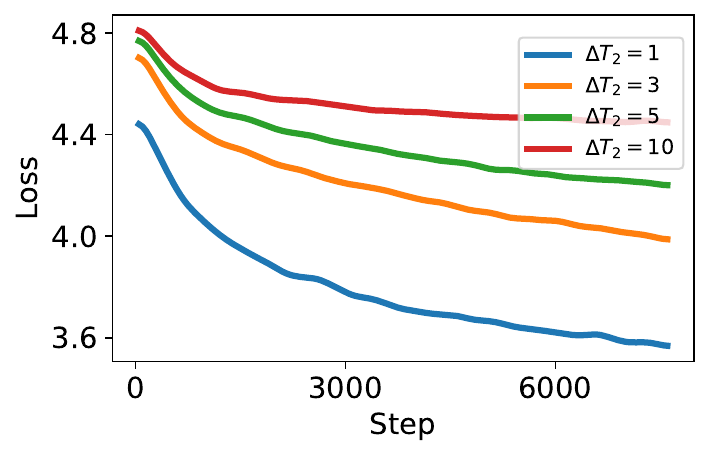}
        \caption{Pretraining loss.} % 如果你不需要caption，可以注释掉这一行
        \label{fig:target window pretraining loss}
    \end{subfigure}
    % \hfill
    \hspace{-0.2cm}
    \begin{subfigure}[b]{0.495\linewidth}
        \includegraphics[width=\textwidth]{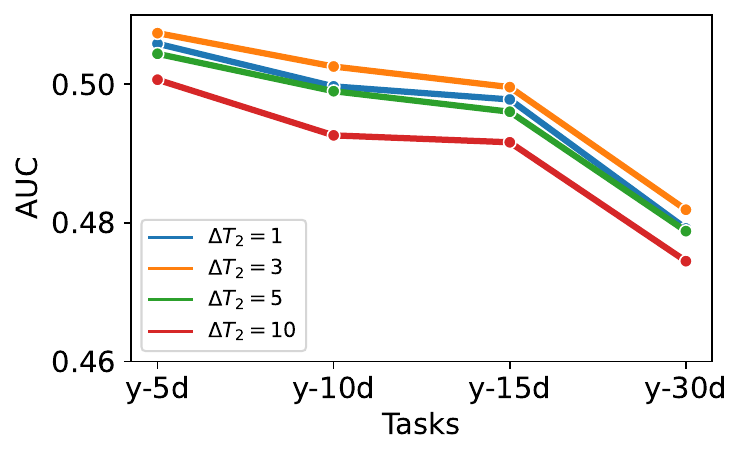}
        \caption{Downstream task AUC.} % 如果你不需要caption，可以注释掉这一行
        \label{fig:target window auc}
    \end{subfigure}
    \vspace{-0.3cm}
    \caption{Prediction window size effect on the Tmall dataset.} % 你可以根据需要添加整个figure的caption
    \label{fig:target pooling window size}
    % \vspace{-.4cm}
\end{figure}

The prediction window size $\Delta T_2$ (rf. Sec. \ref{sec:supervision signal}) is an important hyperparameter in the \model{} pretraining.
Because it determines the supervision signal constructed for pretraining, which directly affects the models' performance.
In this section, we investigate the effect of this hyperparameter for \model{} pretraining.
The experiments are conducted on the Tmall dataset, with prediction window sizes of 1, 3, 5, and 10 days. We mainly investigate the training loss and the finetuning performance on downstream tasks. 
Results are illustrated in Fig. \ref{fig:target pooling window size}.

Fig. \ref{fig:target window pretraining loss} reveals that increasing the prediction window size tends to result in higher training loss. We hypothesize that this is due to the more sophisticated information encompassed in a larger prediction window, making it more challenging to fit compared to a smaller window. 
Nevertheless, a higher training loss does not necessarily lead to inferior downstream performance. 
As shown in Fig. \ref{fig:target window auc}, we can find that $\Delta T_2=1$ and $\Delta T_2=3$ achieve similar performance and are better than $\Delta T_2=10$.
A window size that is too large can negatively impact the quality of the pretrained model.
In practice, we set the prediction window size to one day and it works well for most tasks.
Moreover, the performance of all settings declines as the label becomes far from the current time (5 day$\rightarrow$30 day), which is reasonable since predicting a label more distant from the current observation time is inherently more challenging.
% \xww{revise}

\subsection{Ablation study of different design choices}\label{sec:ablation}

\begin{table}[]
\centering
\caption{Ablation versions and configurations.}
% \vspace{-0.3cm}
\label{tab:abl identifier}
\begin{tabular}{c|ccc}
\hline
Version            & EMA                              & Loss                                                   & Predictor                        \\ \hline
\texttt{Base}      & $\checkmark$                     & CE                                                   & $\checkmark$                     \\
\texttt{EMA}       & \cellcolor[HTML]{D9D9D9}$\times$ & CE                                                  & $\checkmark$                     \\
\texttt{MSE}       & $\checkmark$                     & \cellcolor[HTML]{D9D9D9}MSE                             & $\checkmark$                     \\
\texttt{Predictor} & $\checkmark$                     & CE                                                    & \cellcolor[HTML]{D9D9D9}$\times$ \\ \hline
\end{tabular}
% \vspace{-0.3cm}
\end{table}

% \usepackage[normalem]{ulem}
\begin{table}[]
\centering
\caption{Performance comparison of different ablation versions on the Mobile application dataset.}
% \vspace{-0.3cm}
\label{tab:abl component performance mobile app}
\resizebox{\linewidth}{!}{%
\begin{tabular}{c|cccc}
\hline
                   & Active-30d      & Login-90d      & Unbind-30d     & Unbind-90d      \\ \hline
\texttt{Base}      & 0.9096          & 0.8922         & \textbf{0.668} & {\ul 0.6592}    \\
\texttt{EMA}       & 0.9099          & 0.8900           & 0.6605         & 0.6568          \\
\texttt{MSE}       & \textbf{0.9126} & \textbf{0.895} & {\ul 0.6676}   & \textbf{0.6606} \\
% \texttt{Center}    & {\ul 0.9104}    & {\ul 0.8929}   & 0.6659         & 0.6587          \\
\texttt{Predictor} & 0.5005          & 0.5044         & 0.4915         & 0.4965          \\ \hline
\end{tabular}
}
% \vspace{-0.3cm}
\end{table}

\begin{table}[]
\centering
\caption{Performance comparison of different ablation versions on the Tmall dataset.}
% \vspace{-0.3cm}
\label{tab:abl component performance tmall}
\begin{tabular}{c|cccc}
\hline
                   & 5-day            & 10-day           & 15-day           & 30-day           \\ \hline
\texttt{Base}      & 0.5073          & \textbf{0.5025} & \textbf{0.4996} & {\ul 0.4819}    \\
\texttt{EMA}       & \textbf{0.5082} & {\ul 0.5019}    & 0.4984          & 0.4815          \\
\texttt{MSE}       & 0.4997          & 0.4940          & 0.4900          & 0.4731          \\
\texttt{Predictor} & 0.3556          & 0.3462          & 0.3402          & 0.3207          \\ \hline
\end{tabular}
% \vspace{-0.5cm}
\end{table}

In this part, we investigate the effect of several design choices in the \model{} framework. We primarily focus on the EMA strategy, the loss function, and the predictor module. 
Tab. \ref{tab:abl identifier} lists different version names and configurations.
We compare each version with the \texttt{Base} to investigate the effect of each design.

We present the downstream task performances on both datasets in Tab. \ref{tab:abl component performance mobile app} and Tab. \ref{tab:abl component performance tmall}. The best result is highlighted in bold, and the second best is underlined. 
Removing EMA leads to inferior performance compared to \texttt{Base} in most tasks (6 out of 8 tasks), confirming the necessity of the EMA strategy for stable supervision signals and improved model quality.
Regarding the loss function, both MSE loss and CE loss exhibit strengths on different datasets. The \texttt{MSE} generally outperforms CE (\texttt{Base}) on the Mobile application dataset (3 out of 4 tasks), as shown in Tab. \ref{tab:abl component performance mobile app}. Conversely, CE (\texttt{Base}) performs better on the Tmall dataset, as indicated in Tab. \ref{tab:abl component performance tmall}. We recommend considering both measures for practical applications.
The predictor module is necessary for model pretraining. Without the predictor, the model may not learn well, resulting in worse performance. 
We hypothesize that, without the predictor, the extracted representation may overfit to the supervision embedding, which captures only short-term user behavior. With the predictor, the sequence representation has more flexibility, allowing it to capture richer information about the user and improve performance on diverse downstream tasks.

\subsection{Visualization results}\label{sec: exp visualuzation}

\begin{figure}[ht]
    \centering
    \includegraphics[width=\linewidth]{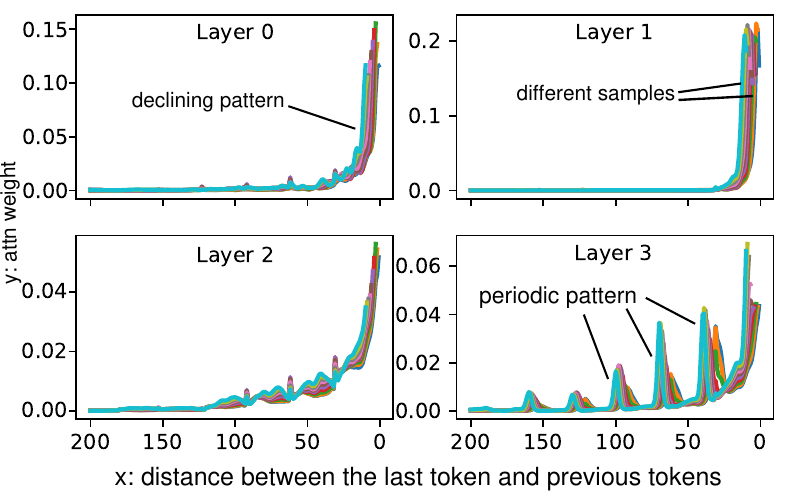}
    \vspace{-0.5cm}
    \caption{Attention patterns of the pretrained \model{}.} % 你可以根据需要添加整个figure的caption
    \label{fig:attention patterns}
    \vspace{-0.2cm}
\end{figure}

\myp{Attention patterns}
To gain a deeper and intuitive understanding of \model{}, we visualize each layer's attention pattern of the pretrained sequential model (\textit{no finetuning}) in Fig. \ref{fig:attention patterns}.
The attention weights are averaged over 512,000 random samples on the Mobile application dataset.
Note that the sequence length is 180 after our behavior pooling (rf. Tab. \ref{tab:dataset statistics}).
We visualize the attention weights of the last ten tokens on their previous tokens.
Note that due to the causal mask, each position only has attention weights with its previous positions. 
The rightmost position of the x-axis in Fig. \ref{fig:attention patterns}, labeled as 0, represents the most recent token position.

In Fig. \ref{fig:attention patterns}, we observe that the model learns to decrease attention weights over time. This is reasonable since historical behaviors distant from the current time likely have diminishing impact on the future. Remarkably, the model also exhibits periodic patterns, particularly in the higher layers, such as layer 2 and layer 3. These periodic attention patterns are inherent to the Mobile Applications dataset, suggesting that considering these periodic user behavior patterns helps improve the prediction of users' future behaviors. \textit{Fig. \ref{fig:attention patterns} demonstrates that \model{} can automatically learn meaningful patterns in the UBS data without any label supervision.}

\begin{figure}[ht]
    \centering
    \begin{subfigure}[b]{0.48\linewidth}
        \includegraphics[width=\textwidth]{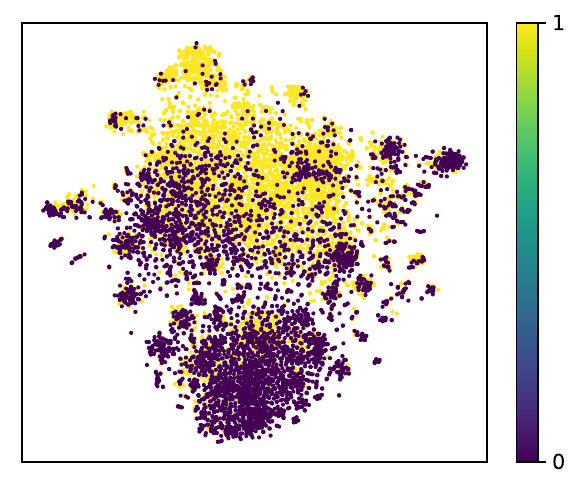}
        \caption{Pretrained representations.} % 如果你不需要caption，可以注释掉这一行
        \label{fig:emb vis before finetune}
    \end{subfigure}
    % \hfill
    \hspace{-0.2cm}
    \begin{subfigure}[b]{0.48\linewidth}
        \includegraphics[width=\textwidth]{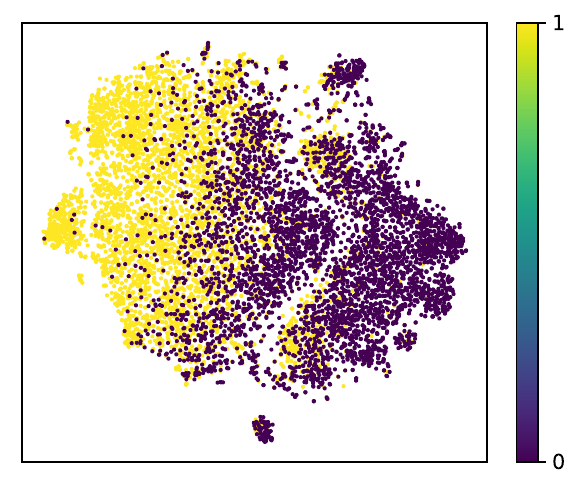}
        \caption{Finetuned representations.} % 如果你不需要caption，可以注释掉这一行
        \label{fig:emb vis after finetune}
    \end{subfigure}
    \vspace{-0.3cm}
    \caption{Representation visualization on the Mobile app dataset (colored-coded via the \textit{Active-30d} label).} % 你可以根据需要添加整个figure的caption
    \label{fig:emb visualization}
    \vspace{-0.2cm}
\end{figure}

\myp{Data representation}
To intuitively investigate the learned representations, we randomly sample 512,000 samples from the Mobile application dataset and extract representations using the pretrained and finetuned \model{} model respectively.
These representations are projected using the t-SNE algorithm \cite{TSNE} and color-coded using the \textit{Active-30d} label.
Fig. \ref{fig:emb vis before finetune} reveals that the pretrained representations \textit{exhibit clustering effects when colored using the downstream label, even though no label was used during pretraining}. This characteristic verifies the effectiveness of the pretraining strategy for downstream classification tasks.
After finetuning, as illustrated in Fig. \ref{fig:emb vis after finetune}, the representations become even more classifiable. This improvement is expected since the label information is utilized during the finetuning stage.

\subsection{Online deployment}\label{sec: exp online exp}
Our model \textbf{has been practically deployed in our industrial production environments for about two months.} The produced user embeddings support two real-world overdue risk prediction tasks for two financial products in Ant group.
The model is pretrained using user behavior data collected from the Alipay mobile application and finetuned using two labels.
We use the Kolmogorov-Smirnov (KS) score to evaluate the model performance.
Experiments indicate that the embeddings could significantly improve the KS by \textbf{about 2.7\% and 7.1\%} for the two business tasks (\textbf{1\% improvement on the KS can already be considered significant, as it can reduce bad debt risk by millions of dollars in Ant group.}) over our baseline model, i.e., the SOTA MSDP method \cite{MSDP}.

\section{Conclusion}
In this paper, we introduce \model{}, a pretraining framework specifically designed for UBS data. Our approach minimizes the need for manual priors and autonomously learns informative representations during pretraining, making it suitable for a variety of downstream tasks. Experimental results indicate that \model{} outperforms current state-of-the-art UBS pretraining methods in both performance and efficiency, while also demonstrating robust scalability in two offline industrial datasets. The \model{} has been deployed in our online production environments for two months and largely improves the KS for two business tasks. In the future, we aim to delve deeper into the intrinsic properties of UBS data and train larger models on more extensive UBS datasets.


% \begin{acks}
% \end{acks}

\bibliographystyle{ACM-Reference-Format}
\bibliography{sample-base}

\appendix

\begin{table*}[t]
\centering
\caption{Comprehensive dataset statistics.}
\label{tab:dataset statistics detailed}
\vspace{-0.2cm}
\resizebox{\linewidth}{!}{%
\begin{tabular}{c|c|ccccccccc}
\hline
Dataset                   & Split    & \# Samples & \# Beh. IDs & Max len & Min len & Avg len & Split window    & Label window    & \# Classes & \# Beh. days \\ \hline
\multirow{3}{*}{Tmall}    & Pretrain & 3.9M       & 8000000     & 600     & 1       & 130.4   & 2013.04-2013.07 & -               & -          & 120          \\
                          & Finetune & 2.6M       & 8000000     & 579     & 1       & 71.6    & 2013.05-2013.06 & 2013.07         & 90         & 60           \\
                          & Test     & 2.9M       & 8000000     & 585     & 1       & 71.3    & 2013.07-2013.08 & 2013.09         & 90         & 60           \\ \hline
\multirow{3}{*}{Mob. app} & Pretrain & 10M        & 30000       & 8994    & 1       & 2129.7  & 2023.03-2023.07 & -               & -          & 180          \\
                          & Finetune & 4.2M       & 30000       & 8990    & 1       & 2114.5  & 2023.07-2023.08 & 2023.09-2023.11 & 2          & 180          \\
                          & Test     & 2.3M       & 30000       & 8991    & 1       & 2104.3  & 2023.10         & 2023.11-2024.01 & 2          & 180          \\ \hline
\end{tabular}
}
\end{table*}

\section{Datasets and preprocessing details} \label{app:datasets}

In this section, we report details of the datasets we utilized for pretraining and downstream finetuning for diverse tasks.
The comprehensive statistics of datasets are shown in Tab. \ref{tab:dataset statistics detailed}.

The Tmall dataset contains about 3.9 million users, 8 million items, and 1.3 billion user-item interactions. The time range spans from 2013.04 to 2013.09. We regard each item as a behavior ID and construct the user behavior sequences by grouping all items a user has interacted with and sorting them chronologically. The dataset contains 90 item categories. 
For downstream tasks, we predict the users' most interested item category (with which the user interacts the most) within a future time window of 5, 10, 15, and 30 days as ground-truth labels.
% (\textit{tasks named 5-day, 10-day, 15-day, and 30-day}).
% The most interested item category is defined as the category with the largest number of items the user will interact with in the defined future time window.
We use four months of data for pretraining, two months of data for finetuning, and one month for testing, as indicated in Tab. \ref{tab:dataset statistics detailed}. For finetuning and testing, we use each user's past 60 days of behaviors.

The mobile app contains about 10 million users. Each user has a behavior sequence recording the click, browsing, and purchase behaviors in the mobile application, with an average sequence length of around 2000.
We utilize a user's past 180 days of behaviors for both pertaining and finetuning.
For downstream tasks, we predict whether a user will be active, login, or unbind his account in the future 30 or 90 days in the mobile application (\textit{tasks named Active-30d, Login-90d, Unbind-30d, and Unbind-90d}).
All tasks are formulated as binary classification problems.
We utilize five months of data for training, two months of data for finetuning, and one month for testing, as indicated in Tab. \ref{tab:dataset statistics detailed}.

\section{Pseudocode for \model{} pretraining}
We illustrate the pseudocode of \model{} pretraining in Algorithm \ref{algo: pretraining}.

\begin{algorithm}
\SetAlgoLined
\KwIn{\textcolor{gray}{student behavior encoder \texttt{SBehEnc}, teacher behavior encoder \texttt{TBehEnc}, sequential model \texttt{SeqModel}, predictor \texttt{Predictor}, behavior pooling window size $\Delta T_1$, prediction window size $\Delta T_2$, temperature $\tau$, momentum $m$, train dataloader \texttt{dataloader}.
}}
\For{{\texttt{(s\_1, s\_2) in dataloader}}}{
    \textcolor{gray}{\# load a batch of data (s\_1, s\_2) with batch size N, s\_1 in the observation window $[0, T]$, s\_2 in the prediction window $[T, T+\Delta T_2]$}\\
    \texttt{e\_s, e\_t = SBehEnc(s\_1), TBehEnc(s\_2)} \textcolor{gray}{\# raw sequence to embeddings with dimension D}\\
    \texttt{e\_s = BehPool(e\_s, $\Delta T_1$)} \textcolor{gray}{\# behavior pooling on dim 1, window size=$\Delta T_1$, stride=$\Delta T_1$, [N, T/$\Delta T_1$, D]}\\
    \texttt{pred = Predictor(SeqModel(e\_s))} \textcolor{gray}{\# extract sequence representation and predict, [N, D]}\\
    \texttt{target = BehPool(e\_t, $\Delta T_2$)} \textcolor{gray}{\# pooling for supervision embedding construction on dim 1, [N, D]}\\
    \texttt{target = target.detach()}
    \textcolor{gray}{\# stop gradient!} \\
    \texttt{loss = loss\_func(pred, target, $\tau$)} \textcolor{gray}{\# compute loss}\\
    \texttt{optimizer.zero\_grad()} \textcolor{gray}{\# clean up gradients } \\
    \texttt{loss.backward()} \textcolor{gray}{\# compute gradients} \\
    \texttt{optimizer.step()} \textcolor{gray}{\# update $\Theta$\_\texttt{SBehEnc}, $\Theta$\_\texttt{SeqModel}, and $\Theta$\_\texttt{Predictor} using gradients}\\
    \texttt{$\Theta$\_TBehEnc = m*$\Theta$\_TBehEnc + (1 - m)*$\Theta$\_SBehEnc} \textcolor{gray}{\# update $\Theta$\_\texttt{TBehEnc} using EMA}
}
\textcolor{gray}{\# loss function}\\
\SetKwFunction{lossFunc}{loss\_func}
\SetKwProg{Fn}{def}{:}{}
\Fn{\lossFunc{\texttt{pred, target, tau}}}{
    \texttt{pred = Softmax(pred/tau, dim=1)} \\
    \texttt{target = Softmax(target/tau, dim=1)} \\
    \texttt{loss = CrossEntropyLoss(pred, target)}\\ 
    \texttt{return loss}
}
\caption{PyTorch-like pseudocode for \model{} pretraining}
\textcolor{gray}{\# Note: By default, we use the cross entropy loss.}
\label{algo: pretraining}
\end{algorithm}

\section{Details of the scaled models}\label{sec: appendix scaling effect}

\begin{table}[]
\centering
\caption{Configuration of the sequential model for different model sizes.}
\label{tab:scaling model config}
\begin{tabular}{c|ccccc}
\hline
                        & Dim & FF Dim       & Layer & \# Params & Scale factor \\ \hline
\rowcolor[HTML]{EFEFEF} 
\texttt{Base}           & 128 & 128          & 4     & 395264 & $\times$1    \\
\texttt{Base}$\times2$  & 128 & 128          & 8     & 790528 & $\times$2    \\
\texttt{Base}$\times4$  & 128 & 256          & 5     & 1643520 & $\times$4   \\
\texttt{Base}$\times8$  & 128 & 256          & 10    & 3287040 & $\times$8   \\
\texttt{Base}$\times16$ & 128 & {512} & 5     & 5908480 & $\times$16   \\ \hline
\end{tabular}
\end{table}

We list the configurations of different scaled models in Tab. \ref{tab:scaling model config}. 
Dim indicates the output dimension of the transformer layer, and FF Dim represents the feed-forward hidden dimension in the feedforward layers of the transformer encoder layer.
Please note that we approximate the model parameters to be multiples of those in the base version, as extensively exploring architectural choices is impractical during training. For all models, we utilize identical configurations for the behavior encoder, the embedding table, and the prediction heads in the downstream finetuning stage. During finetuning, all models' encoders are frozen, and only prediction heads are finetuned.
The \texttt{Base} model in Tab. \ref{tab:scaling model config} is identical to the \model{}(freeze) model used in Tab. \ref{tab:main results}.

\end{document}